\documentclass[sigconf]{acmart}
\usepackage{algorithm,algorithmic}
\usepackage{amsmath,amsthm}
\usepackage{subcaption}

\AtBeginDocument{%
  }

\begin{document}

\setcopyright{acmlicensed}
\copyrightyear{2025}
\acmYear{2025}
\setcopyright{acmlicensed}\acmConference[KDD '25]{Proceedings of the 31st ACM SIGKDD Conference on Knowledge Discovery and Data Mining V.2}{August 3--7, 2025}{Toronto, ON, Canada}
\acmBooktitle{Proceedings of the 31st ACM SIGKDD Conference on Knowledge Discovery and Data Mining V.2 (KDD '25), August 3--7, 2025, Toronto, ON, Canada}
\acmDOI{10.1145/3711896.3737241}
\acmISBN{979-8-4007-1454-2/2025/08}


\title{Lightweight Auto-bidding based on Traffic Prediction in \\ Live Advertising}

\author{Bo Yang}
\authornote{Both authors contributed equally to this research.}
\email{yb342827@taobao.com}
\affiliation{%
  \institution{Taobao \& Tmall Group of Alibaba}
  \state{Beijing}
  \country{China}
}

\author{Ruixuan Luo}
\email{luoruixuan.lrx@taobao.com}
\authornotemark[1]
\affiliation{%
  \institution{Taobao \& Tmall Group of Alibaba}
  \state{Beijing}
  \country{China}
}

\author{Junqi Jin}
\email{junqi.jjq@taobao.com}
\authornote{Junqi Jin is the corresponding author.}
\affiliation{%
  \institution{Taobao \& Tmall Group of Alibaba}
  \state{Beijing}
  \country{China}
}

\author{Han Zhu}
\email{zhuhan.zh@taobao.com}
\affiliation{%
  \institution{Taobao \& Tmall Group of Alibaba}
  \state{Beijing}
  \country{China}
}

\renewcommand{\shortauthors}{Bo Yang, Ruixuan Luo, Junqi Jin, \& Han Zhu}




\begin{abstract}
Internet live streaming is widely used in online entertainment and e-commerce, where live advertising is an important marketing tool for anchors. An advertising campaign hopes to maximize the effect (such as conversions) under constraints (such as budget and cost-per-click). The mainstream control of campaigns is auto-bidding, where the performance depends on the decision of the bidding algorithm in each request. The most widely used auto-bidding algorithms include Proportional–Integral–Derivative (PID) control, linear programming (LP), reinforcement learning (RL), etc. Existing methods either do not consider the entire time traffic, or have too high computational complexity. In this paper, the live advertising has high requirements for real-time bidding (second-level control) and faces the difficulty of unknown future traffic. Therefore, we propose a lightweight bidding algorithm Binary Constrained Bidding (BiCB), which neatly combines the optimal bidding formula given by mathematical analysis and the statistical method of future traffic estimation, and obtains good approximation to the optimal result through a low complexity solution. In addition, we complement the form of upper and lower bound constraints for traditional auto-bidding modeling and give theoretical analysis of BiCB. Sufficient offline and online experiments prove BiCB's good performance and low engineering cost.
\end{abstract}

\ccsdesc[500]{Information systems~Display advertising}
\ccsdesc[500]{Applied computing~Electronic commerce}

\keywords{Live Streaming, Advertising, Auto-bidding, Traffic Prediction}


\maketitle

\newcommand\kddavailabilityurl{https://doi.org/10.5281/zenodo.15552885}

\ifdefempty{\kddavailabilityurl}{}{
\begingroup\small\noindent\raggedright\textbf{KDD Availability Link:}\\
The source code of this paper has been made publicly available at \url{\kddavailabilityurl}.
\endgroup
}

\section{Introduction}
\label{intro}

Internet live streaming has become the mainstream form of media and e-commerce \cite{yu2023curriculum}. In order to actively reach consumers to increase popularity and sales, live advertising is an important marketing tool for anchors. Similar to traditional image advertising, live advertising generally adopts a real-time bidding mechanism (RTB) \cite{zhang2014optimal, wen2022cooperative,ou2023deep,zhu2017optimized}. In RTB, the campaign created by the advertiser generally requires maximizing the effect (such as conversions and number of fans) under constraints (such as budget and cost-per-click) \cite{guo2021we}. When a consumer requests a live advertising system, the platform sorts each ad with a bid based on an auction mechanism, such as Generalized Second Price (GSP) \cite{edelman2007internet,aggarwal2009general}, and the ad with the highest bid wins the display opportunity and is charged. The mainstream advertising campaign control is based on auto-bidding, which combines the basic information of the campaign and the current situation of traffic to calculate the bidding decision for each request. Unlike traditional advertising, live advertising has high requirements for timeliness. As the anchor tells the climax and trough of the content, the anchor hopes that the live advertising can detonate the live room within a few minutes, which requires real-time perception and control of advertising at the second level.

\begin{figure*}
    \centering
    \includegraphics[width=0.95\linewidth]{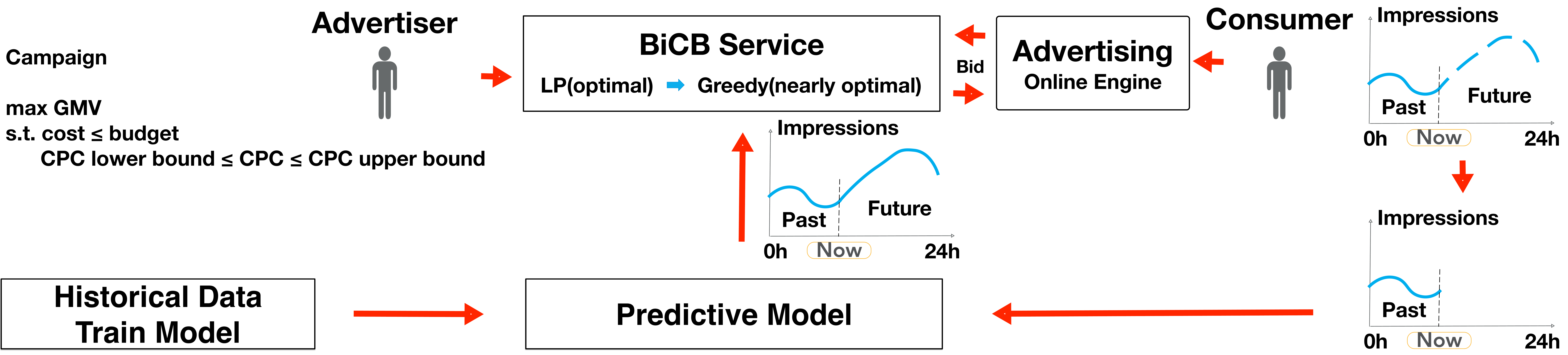}
    \caption{Overview of BiCB auto-bidding system: An advertiser's campaign hopes to maximize gross merchandise volume (GMV) under constraints such as cost-per-click (CPC) within upper bound and lower bound. Consumers' requests until now as traffic are fed to a Predictive Model to estimate entire time traffic. BiCB uses all above information to compute bid decision with a nearly optimal greedy algorithm.}
    \label{fig:intro-overview}
\end{figure*}


%
There is a consensus mathematical modeling for auto-bidding from the advertiser's perspective, that is, a linear programming problem that maximizes a linear objective and satisfies multiple linear constraints. The objective function is the sum of the values of all traffic requests. The problem can be solved optimally by an LP solver, however, this requires knowing all the traffic in advance. In practice, at any time, a campaign needs to make a bidding decision for the current request, but at this time, the algorithm cannot know the future traffic and cannot change historical decisions. This is the biggest challenge for auto-bidding.

The main solutions in practice include: PID control \cite{yang2019bid,johnson2005pid}, linear programming (LP) \cite{schrijver1998theory}, reinforcement learning (RL) \cite{cai2017real,he2021unified,mou2022sustainable,zhang2023personalized,hao2020dynamic,zhang2023personalized,DBLP:conf/cikm/WuCYWTZXG18}, and recently proposed artificial intelligence generative bidding (AIGB) \cite{guo2024generative}. PID control collects the effect data of a recent period of time (e.g. last 5 minutes) and examines whether the data of this small period violates the constraints, thereby calculating the error of the local time period based on PID and adjusting the bidding decision. PID's engineering is simple and has good interpretability, but its bidding decision only relies on local time information and does not consider full-time information, so it is suboptimal.

Linear programming (LP) takes yesterday's full-time traffic details as input and calls a professional LP solver to find the optimal bid at each moment. For today's campaign, the bid decision at the same moment yesterday is queried. Another variant is that for today's campaign, yesterday's detailed data between today's now and the end of the campaign is input into the LP solver to obtain the optimal bid. The defect of LP is that it relies on detailed data rather than data distribution, which has poor generalizability, and the campaign delivery time setting may be different, resulting in large errors. The LP solution has high computational complexity, and in practice, it is necessary to downsample the data by a large proportion, which also reduces the performance.

RL and AIGB regard auto-bidding over time as a sequential decision problem. RL learns the full-time sequential decision of historical campaigns, and designs the objective function and constraints as rewards, thereby learning a sequential bidding agent. AIGB learns the correlation between the full-time bidding sequence and reward of the historical campaigns, and summarizes a function that can generate future bidding sequences based on historical bidding sequences. These two methods can optimize full-time bidding, and the models based on data distribution have good generalizability. However, in practice, sequence modeling requires exploring more samples, which will cause economic losses when deployed online, and deployment in an offline simulation environment requires solving the inconsistency problem between offline and online environments. Due to the large sample demand of sequence modeling, training is more difficult to converge than supervised learning.

After re-examining the linear characteristics of auto-bidding modeling, we think that sequence modeling is not necessary. We consider the simplest modeling of maximizing conversions under a budget constraint bidding (BCB), with a linear objective function and a linear constraint. The intuitive solution is linear programming. In addition, the BCB is a typical knapsack problem, which can be solved by a greedy algorithm, that is, all traffic requests are sorted according to the cost-effectiveness of conversions and cost, and request with high cost-effectiveness is prioritized, similar to the priority of good items entering the bag, until the accumulated cost of traffic (items) is equal to the budget (capacity of the bag), an approximate optimal solution can be obtained, and the granularity of a single traffic request is small enough, which means the optimal approximation of the greedy algorithm is very good. It should be noted that in this algorithm, the order in which items enter the bag is not important. We only need to find the cost-effectiveness threshold for items to enter the bag, and then it is feasible for items to enter the bag in any random order. The search for the cost-effectiveness threshold only depends on the understanding of the cost and conversions distribution of all items, and does not depend on a certain sequence process. The core condition that needs to be solved here is to know the full-time traffic in advance.

Thus, we propose a method consisting of two parts: first, predicting the full-time traffic of a campaign. We do not use sequence modeling, nor do we estimate future traffic details. Therefore, the representation form, estimation method, and good generalizability of full-time traffic are full of challenges; second, based on the estimation results, design a solution to approach the LP optimal result. How the algorithm can avoid high computational complexity and how to use full-time traffic estimation are also full of challenges.

To address the above challenges, we propose BiCB algorithm as Figure \ref{fig:intro-overview}. We improve the traditional modeling that only considers upper bound constraints and supplement the modeling of upper and lower bound constraints. We regard this problem as a generalized knapsack problem and design a lightweight method similar to the greedy algorithm. Specifically, we perform Lagrangian dual analysis \cite{frangioni2005lagrangian} on the original problem and obtain the judgment criteria for whether a request should be won. The calculation of this criteria depends on the optimized values of the dual variables. When making online decisions, the request is calculated according to the judgment criteria in turn. If it meets the criteria, it will be won, and if it does not meet the criteria, it will be abandoned. We prove that this method has excellent approximation ability for the LP optimal solution. The dual variable needs to be solved based on the estimation of the full-time traffic. The general idea is to try a set of dual variables, execute the full-time traffic based on the judgment criteria calculated by dual variables, and observe whether each constraint is finally satisfied. If not, adjust the values of the dual variables and try again until the constraint is satisfied, and then we obtain the optimal dual variables. The execution results of all traffic corresponding to a set of dual variables are obtained by calling the estimation of full-time traffic. We train an estimation model for the dual variables to the execution results of time segments based on historical data. This model only needs to estimate the cumulative value of a time segment, such as cumulative cost and cumulative clicks, and does not need to estimate the details of traffic, so it can significantly improve the accuracy of the estimation. We use the data of all advertisers from multiple days in history to train this estimation model, which significantly improves the generalizability.

The contributions of this paper mainly include: 1) Based on the existing modeling that only considers the upper bound constraints, we provide a more complete modeling that includes upper and lower bounds; 2) We propose a lightweight bidding algorithm BiCB based on the prediction of full-time traffic; 3) We give theoretical analysis of the BiCB method, which has excellent approximate performance for optimal solution; 4) We give an engineering implementation overview for live streaming advertising; 5) We discuss the pros and cons of different mainstream methods of auto-bidding; 6) BiCB has been fully deployed online and achieved good business results.

\section{Formulation and Dual Analysis}

Recent works, such as BCB\cite{DBLP:conf/cikm/WuCYWTZXG18}, Multi-Constrained Bidding (MCB) \cite{yang2019bid}, and Unified Solution to Constrained Bidding (USCB)\cite{he2021unified}, provide problem formulation under budget constraints and cost-per-click (CPC) or KPI upper bound constraints, but do not consider lower bound constraints of CPC or KPIs. In order to screen the quality of traffic, advertisers sometimes hope that the CPC, CTR, ROI, etc. of the request are greater than a lower bound. In this section, we extend the problem formulation with lower bound constraints based on existing works and discuss it using the most common CPC lower bound constraint as an example.

Assume that an advertiser can bid on $N$ impressions in a day, the click-through rate of winning the i-th impression is $pctr_i$, the CPC of winning the i-th impression is $wp_i$, and the value brought to the advertiser after click (such as conversion rate) is $obj_i$. Assume that the advertiser's budget is $B$, and the upper and lower bounds of the CPC are $C_u$ and $C_l$ respectively. Then the problem can be formulated as the following linear programming problem:

\begin{align}
\mathop{max}\limits_{x_i}&&{C_u\sum_i x_i*pctr_i*obj_i} \label{eq:lp1} \tag{LP1} \\
s.t. &&~\sum_i x_i*pctr_i*wp_i \le B \\
&&\frac{\sum_i x_i*pctr_i*wp_i}{\sum_i x_i*pctr_i} \le C_{u} \\
&&\frac{\sum_i x_i*pctr_i*wp_i}{\sum_i x_i*pctr_i} \ge C_{l} \\
&&0\le x_i \le 1,\forall i
\end{align}

Here, $x_i$ indicates whether the advertiser can win the traffic. In reality, $x_i$ should only have two states: 0 and 1. However, for convenience in solving, $x_i$ is defined as a continuous value between 0 and 1 as in \cite{yang2019bid, he2021unified}. In the following text, we will also prove that the error in the optimal solution of this linear programming formulation and the real 0-1 programming formulation is controllable.

The dual problem of (\ref{eq:lp1}) can be written as:

\begin{align}
\mathop{min}\limits_{p, q_u, q_l,r_i}&&B*p+\sum_i r_i \\
\nonumber
s.t.&&wp_i*pctr_i*p+(wp_i-C_u)*pctr_i*q_u \\
&&-(wp_i-C_l)*pctr_i*q_l+r_i \ge C_uv_i,~\forall i \label{eq:lp_con0} \\
&&p,q_u,q_l \ge 0 \\
&&r_i \ge 0,~\forall i \\
&&v_i = pctr_i*obj_i,~\forall i
\end{align}

According to the complementary relaxation theorem, the necessary and sufficient conditions for the feasible solutions $x_i^*$, $p^*$, $q_u^*$, $q_l^*$, $r_i^*$ to be optimal solutions are as follows:

\begin{align}
\nonumber
x_i^**((\frac{C_u*obj_i+C_u*q_u^*-C_l*q_l^*}{p
^*+q_u^*-q_l^*}-wp_i) \\
*pctr_i*(p^*+q_u^*-q_l^*)-r_i^*)=0,~\forall i \label{eq:relax_x0}\\
p^**(\sum_i x_i^**pctr_i*wp_i - B)=0 \label{eq:relax_B}\\
q_u^**(\frac{\sum_i x_i^**pctr_i*wp_i}{\sum_i x_i^**pctr_i} - C_{u})=0 \label{eq:relax_cu}\\
q_l^**(\frac{\sum_i x_i^**pctr_i*wp_i}{\sum_i x_i^**pctr_i} - C_{l})=0 \label{eq:relax_cl}\\
(x_i^*-1)*r_i^* = 0,~\forall i \label{eq:relax_x1}
\end{align}

We discuss in two cases $p^*+q_u^*-q_l^*>0$ and $p^*+q_u^*-q_l^* \le 0$.

\begin{itemize}
    \item $p^*+q_u^*-q_l^*>0$

    Define $bid_i^*=\frac{C_u*obj_i+C_u*q_u^*-C_l*q_l^*}{p
^*+q_u^*-q_l^*}$.

\begin{itemize}
    \item If the advertiser wins the i-th impression, we have $x_i^*>0$, combining (\ref{eq:relax_x0}) we can get $bid_i^*\ge wp_i$.

    \item If the advertiser does not win the i-th impression, we have $x_i^*=0$, combining (\ref{eq:relax_x1}) we can get $r_i^*=0$, and according to (\ref{eq:lp_con0}) we can obtain $bid_i^*\le wp_i$.

\end{itemize}
    Therefore, when $p^*+q_u^*-q_l^*>0$, the optimal solution is bidding with equation $bid_i^*=\frac{C_uobj_i+C_u*q_u^*-C_l*q_l^*}{p
^*+q_u^*-q_l^*}$.

    \item $p^*+q_u^*-q_l^* \le 0$

Since at most one of $\frac{\sum_i x_i^**pctr_i*wp_i}{\sum_i x_i^**pctr_i}=C_{u}$ and $\frac{\sum_i x_i^**pctr_i*wp_i}{\sum_i x_i^**pctr_i}\\ = C_{l}$ satisfies, according to (\ref{eq:relax_cu}) and (\ref{eq:relax_cl}), at most one of $q_u^*$ and $q_l^*$ is non-zero, so $q_u^*=0$. In this situlation, only the CPC lower bound restriction is effective.

\begin{itemize}
    \item If the advertiser wins the i-th impression, we have $x_i^*>0$, according to (\ref{eq:relax_x0}) we can get $C_u*obj_i-C_l*q_l^*\ge wp_i*(p^*-q_l^*)$.

    \item If the advertiser does not win the i-th impression, we have $x_i^*=0$, according to (\ref{eq:relax_x1}) and (\ref{eq:lp_con0}) we can obtain $C_u*obj_i-C_l*q_l^*\le wp_i*(p^*-q_l^*)$.

\end{itemize}
    Therefore, when $p^*+q_u^*-q_l^*=0$, the optimal solution is winning all the impressions that satisfies $obj_i\ge \frac{C_l}{C_u}*q_l^*$.

    When $p^*+q_u^*-q_l^*<0$, the optimal solution is winning all the impressions that satisfies $\frac{C_u*obj_i-C_l*q_l^*}{p^*-q_l^*}\le wp_i$.
    
\end{itemize}

From the above discussion, it can be seen that after introducing the lower bound constraint on CPC, the optimal solution to this problem is not necessarily equivalent to an auction system. If $p^*+q_u^*-q_l^*\le 0$ is satisfied, in order to satisfy the lower bound constraint, it is necessary to deliberately take some high-priced and low-value impression. However, in real scenarios, the value of an impression to all bidders is not independent of each other. For all advertisers, the conversion rate of a user with high purchasing power will be significantly higher than that of a user with low purchasing power. Therefore, there is often an obvious positive correlation between the value of impression and the cost required to win it. We have also proved that when there is a high positive correlation between value and cost, the optimal solution satisfies $p^*+q_u^*-q_l^*>0$, that is, the optimal solution still satisfies the equivalence of an auction system. The proof will be shown in the appendix. 

\section{Solution based on Traffic Prediction}

The above discussion shows that based on the acquisition of all the traffic for the whole day in advance, the optimal dual variable can be solved through linear programming. In a real-world scenario, the following steps are required to achieve the theoretical optimal solution:

\begin{enumerate}
    \item Build a Time Machine.
    \item Get detailed data of all-day impression through the time machine.
    \item Based on the above optimization problem, calculate the optimal dual variable.
    \item Go back to the starting point of the day, bid according to the solved dual variable, and execute the optimal solution throughout the day.
\end{enumerate}

Therefore, in order to achieve the optimal solution, the following capabilities are required:

\begin{itemize}
    \item Time machine capability: Need to be able to accurately predict the future impressions throughout the day.
    \item Optimal bidding formula: Based on the analytical conclusions of mathematical modeling, solve the optimal dual variable and bid with a fixed optimal formula throughout the day.
    \item A fast algorithm: Due to the high computational complexity of linear programming, a more efficient algorithm is needed online to approximate the optimal solution.
\end{itemize}

According to equations (\ref{eq:relax_B}), (\ref{eq:relax_cu}), and (\ref{eq:relax_cl}) , the necessary and sufficient conditions for the optimal solution are:

\begin{itemize}
    \item $p^*=0$, or $p^*>0$ and the total cost is $B$.
    \item $q_u^*=0$, or $q_u^*>0$ and the total CPC is $C_u$.
    \item $q_l^*=0$, or $q_l^*>0$ and the total CPC is $C_l$.
\end{itemize}

Therefore, our "time machine" only needs to know the total cost and CPC for the whole day when bidding with the dual variables of $p, q_u, q_l$, and then it can determine whether the current solution is the optimal solution.

Based on this analysis, we propose a bidding algorithm BiCB based on future traffic estimation, which is a greedy algorithm. First, in order to obtain the "time machine", a regression model (LightGBM \cite{ke2017lightgbm} is used in this paper) is trained through historical data to fit the functions $C(t, features, p, q_u, q_l)$ and $K(t, features, p, q_u, q_l)$, which represent the cost and click in time period $t$ with dual variables $p$, $q_u$, and $q_l$, respectively.
At time $t_0$, if we bid with parameter $p$, $q_u$ and $q_l$ in the remaining time, the advertiser's cumulative cost and click for the whole day can be expressed as:

\begin{align}
COST(t_0,p,q_u,q_l)=acc\_cost(t_0)+\sum_{t=t_0}^{t_{end}}C(t,features,p,q_u,q_l) \\
CLK(t_0,p,q_u,q_l)=acc\_clk(t_0)+\sum_{t=t_0}^{t_{end}}K(t,features,p,q_u,q_l)
\end{align}

$acc\_cost(t_0),acc\_clk(t_0)$ respectively represent the cost and click generated from the start of the campaign to $t_0$.

In the actual online delivery process, we use the projected gradient descent algorithm to adjust the $p$, $q_u$, and $q_l$ parameters in real time to satisfy equations (\ref{eq:relax_B}), (\ref{eq:relax_cu}), and (\ref{eq:relax_cl}). The specific algorithm flow is shown in Algo. \ref{algo1}:

\begin{algorithm}
\caption{Algorithm BiCB}
\begin{algorithmic}
\REQUIRE $B$: Budget, $C_u$: CPC upper bound, $C_l$: CPC lower bound, $t_0$: current time
\ENSURE $p,q_u,q_l$: optimal parameters
\STATE load model parameters
\STATE initialize $p,q_u,q_l$ with current parameters
\WHILE{$p,q_u,q_l$ not converged}
    \STATE $pCOST \leftarrow COST(t_0,p,q_u,q_l)$
    \STATE $pCLK \leftarrow CLK(t_0,p,q_u,q_l)$
    \STATE $L_p \leftarrow (B-pCOST)^2$
    \STATE $L_{qu} \leftarrow (C_u*pCLK-pCOST)^2$
    \STATE $L_{ql} \leftarrow (C_l*pCLK-pCOST)^2$
    \STATE calculate step length $\alpha_i,\beta_i,\gamma_i$
    \STATE $p \leftarrow max(0,p-\alpha_i*\frac{\partial L_p}{\partial p})$
    \STATE $q_u \leftarrow max(0,q_u-\beta_i*\frac{\partial L_{qu}}{\partial q_u)}$
    \STATE $q_l \leftarrow max(0,q_l-\gamma_i*\frac{\partial L_{ql}}{\partial q_l)}$
    \STATE $i \leftarrow i+1$
\ENDWHILE
\RETURN $p,q_u,q_l$
\end{algorithmic}
\label{algo1}
\end{algorithm}

In order to find the optimal $p, q_u, q_l$ parameters, we use the projected gradient descent method \cite{chen2019non} to find the optimal solution. It can be proved that the the direction of the optimal solution for all the dual variables are opposite to the direction of the gradient. Further, it can be proved that when the step length is set reasonably, for any $\epsilon>0$, this method can converge to the optimal solution in no more than $O((log\frac{1}{\epsilon})^2) $ iterations. For detailed proof, please refer to the appendix.

\section{Theoretical Analysis}

We have proved that for the defined linear programming problem, the optimal solution is to bid according to  the equation $bid_i^*=\frac{C_uobj_i+C_u*q_u^*-C_l*q_l^*}{p
^*+q_u^*-q_l^*}$. In this section, we will discuss the error between the linear programming problem formulation and the real online 0-1 programming scenario, and prove that the error between the greedy solution of bidding according to the fixed formula and the optimal solution of the 0-1 programming problem is controllable.

For a problem instance $(wp, obj, B, C_u, C_l), wp>0, obj>0, B>0, C_u>C_l>0$, we define a greedy solution algorithm for the 0-1 programming problem as:

\begin{enumerate}
    \item Calculate the optimal dual variables $p^*, q_u^*, q_l^*$ of the linear programming problem $(wp, obj, B, C_u-\epsilon^c, C_l+\epsilon^c)$ 
    \item If $obj_i>\frac{wp_i*(p^*+q_u^*-q_l^*)-C_u*q_u^*+C_l*q_l^*}{C_u}$, then set $x_i=1$, otherwise $x_i=0$
\end{enumerate}

$\epsilon^c$ is used to make sure that the greedy algorithm satisfies the CPC constraint, which is defined below.

We define the optimal solution and greedy solution of the 0-1 programming problem and the linear programming problem as follows:
\begin{itemize}
    \item The optimal solution of the 0-1 programming problem is $V_d^*(wp,obj,B,C_u,C_l)$
    \item The greedy solution of the 0-1 programming problem is $V_d^G(wp,obj,B,C_u,C_l)$
    \item The optimal solution of the linear programming problem is $V_c^*(wp,obj,B,C_u,C_l)$
\end{itemize}

\newtheorem{thm1}{Theorem}

\begin{thm1}
\label{thm1}
$V_d^G(wp,obj,B,C_u+\epsilon^c,C_l-\epsilon^c)\ge V_d^*(wp,obj,B,C_u\\,C_l)-2\mathop{max}\limits_i obj_i$, where $\epsilon^c=max(\frac{2C_u^2}{B-2C_u},\frac{2C_l\mathop{max}\limits_i wp_i}{B})$
\end{thm1}

Theorem \ref{thm1} proves that the error of the total value and CPC between the greedy solution and the optimal solution of the online 0-1 programming problem is controllable. For BCB problem, it is equivalent to the knapsack problem, and can be easily proved that when the granularity of items is small enough, the greedy solution is almost equivalent to the optimal solution. And for BiCB problem, we have similar proof. The detailed proof of the theorem will be shown in the appendix. Here we give an example of the theorem, for an actual online advertising campaign, if the total budget is not less than 300 yuan, the upper and lower bounds of the cost-per-click are 1.2 yuan to 2 yuan, the highest bid for a single click does not exceed 3 yuan, the total transaction volume is not less than 100, and the conversion rate of a single click is not higher than 1, it can be deduced that the error between the greedy solution and the optimal solution does not exceed 2\%. 

\section{Implementation} 

The implementation of the algorithm pipeline is shown in Figure \ref{fig:implementation}:

\begin{figure*}
    \centering
    \includegraphics[width=\linewidth]{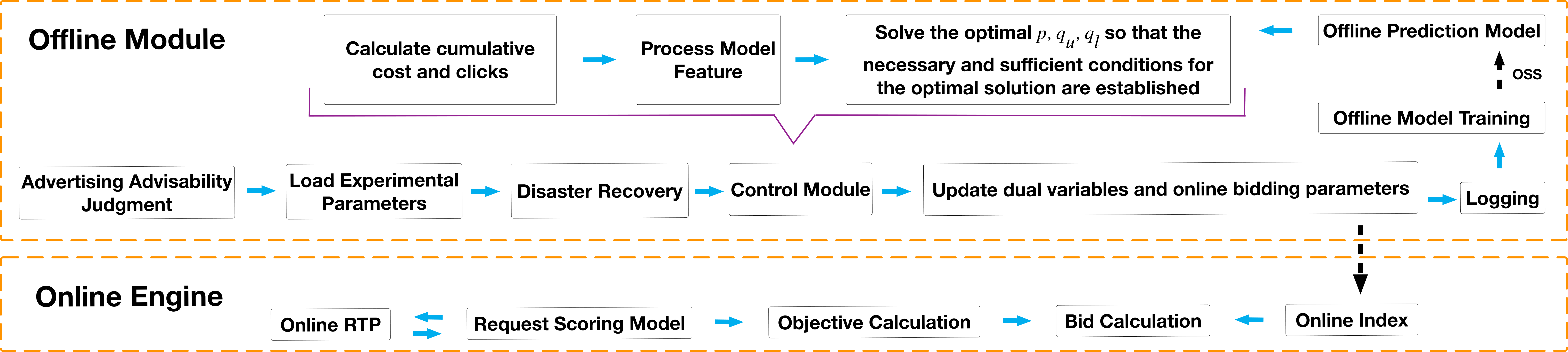}
    \caption{Online and offline engineering implementation.}
    \label{fig:implementation}
\end{figure*}
        
        A flink-based \cite{carbone2015apache} offline control module is deployed to perform offline control every 10 seconds to ensure real-time performance in live advertising. The module consists of several stages:
        \begin{enumerate}
            \item Advertising advisability judgment: Determine whether a campaign can be launched based on its launch schedule.
            \item Loading experimental parameters: Load experimental parameters for campaign experiments or budget experiments. Our offline system supports campaign experiments and budget experiments. In campaign experiments, different experiment parameters can be configured for different campaigns. In a budget experiment, the budget of each campaign is divided into several equal parts, each of which can be configured with different experimental parameters and regulated independently.
            \item Disaster recovery: Monitor some of the potential risks in the control process (e.g., data delay, etc.) to prevent bad advertiser experience by spending budget too quickly or having a high CPC. When a potential risk is detected, the parameters need to be frozen or reset in real time.
            \item Statistics on real-time cost and clicks: Calculate cumulative cost and clicks of each campaign in flink real-time data stream and use "pre-charge" technic to calibrate the data to overcome click data delay problem, thereby obtaining more accurate real-time data.
            \item Offline control: Construct features of current campaign required by offline model according to real-time cost and CPC, then load offline model from OSS (Object Storage Service) and calculate optimal dual variables.
            \item Update of Results: Update the values of the dual variables and the online bidding parameters according to the optimal solution.
            \item Logging: Write current used dual variables, cumulative cost, clicks and some other information into the offline algorithm log for subsequent model training.
        \end{enumerate}
        
        In online engine, the primary stages of bidding module are as follows:
        \begin{enumerate}
            \item Requesting RTP: The online engine will request another RTP (Realtime Prediction) service to estimate various scores (e.g. CTR , CVR) with a DNN model \cite{sheng2024enhancing,zhou2018deep} for each campaign. 
            \item Objective calculation: Calculate the objective value according to the scores provided by model and the type of the campaign (e.g., for campaigns that aim to maximize conversions, the objective value is the conversion rate; for campaigns that aims to gain followers, the objective value is the follower growth rate). 
            \item Bid calculation: In the design of the entire advertising system, the offline module is primarily responsible for business logic, and the online module is mainly responsible for functional calculations. In order to avoid the need to modify the online bidding formula every time a new bidding algorithm is proposed, our online module abstracts the final bidding formula into $bid=\alpha*obj+\beta$. Different bidding algorithms only need to calculate the corresponding values of $\alpha$ and $\beta$, so there is no need to modify the online code when launching a new bidding algorithm. For BiCB algorithm, $\alpha=\frac{C_u}{p^*+q_u^*-q_l^*},~\beta=\frac{C_u*q_u^*-C_l*q_l^*}{p^*+q_u^*-q_l^*}$. After getting an optimal solution, the offline module calculates $\alpha$, $\beta$, writes them into the online index, and the online engine then reads the index and uniformly uses this formula for bidding.
        \end{enumerate}
        
        We use our real online delivery data to train offline prediction model. Our training procedure is as follows:
        \begin{enumerate}
            \item At the end of the day, samples of offline algorithm log within the past week are collected as training set.
            \item Collect bidding parameters, actual cost, and clicks in each minute as training samples. And then train the model.
            \item Synchronize the updated model to OSS for use by the offline control module.
        \end{enumerate}

\section{Discussion}
Finally, we discuss the technical comparison of several best methods and our method. Yang \cite{yang2019bid} first proposed multi-constrained bidding modeling (MCB) based on dual PID control. This work provides a dual analysis method that is widely recognized at present. Its dual PID control is also simple and clever, and is widely used in industry. Its shortcomings are that the control only relies on local time window information, there is no full-time optimization, and its dual PID algorithm does not provide theoretical analysis. Our work improves these shortcomings.

Cai \cite{cai2017real} proposed an RL solution modeling bidding in 2017. RL mainly considers the state transition of remaining budget and traffic features. The paper points out that the training of RL value function is complex. He \cite{he2021unified} first proposed a bid modeling scheme with multiple upper bound constraints and used RL to solve it. Based on this work, we supplement the bid modeling scheme with upper and lower bound constraints to make the bidding modeling complete. Mou \cite{mou2022sustainable} pointed out that the online exploration economic cost of RL modeling bidding is high, but the use of offline simulation environment has environmental errors. Therefore, an optimization solution for online exploration is given to accelerate training convergence. The RL modeling proposed by Hao \cite{hao2020dynamic} is used for sequential reach on the consumer side, rather than for advertising campaign decisions for all-day traffic. When optimizing for all-day traffic, the paper uses the simple greedy knapsack problem algorithm instead of RL modeling. Zhang \cite{zhang2023personalized} based on RL modeling learning agent to improve the fairness of platform agent bidding.

Early RL modeling \cite{kaelbling1996reinforcement} completely abandoned the mathematical modeling of LP, making the bid directly equal to the output of the black box neural network. The huge search space of the bid made it difficult for training to converge. Later RL modeling first reuses the conclusion of LP dual analysis, and then uses RL methods to solve the dual variables. We think that LP modeling of auto-bidding is similar to the knapsack problem. The key to solving it is the advance knowledge of all items and the overall distribution of the value cost of the items. The order in which the items are selected is not important, so sequence modeling is not necessary. In addition, RL methods usually use dual variables as actions and sequence cumulative value (i.e., cumulative conversions of advertising campaigns) as rewards for modeling. Accordingly, the value function in RL needs to estimate the future cumulative conversions based on the current dual variables, so that the value function is called during inference to obtain the action (dual variable) that maximizes the cumulative conversions. In our method, we only need to estimate the future cumulative cost and click based on the current dual variables. In cost-per-click (CPC) advertising systems, consumer events occur in the order of exposure, clicks, cost, and conversions. Conversion data is much sparser than click and cost data and has a greater time delay, which makes the estimation of cumulative conversions a more difficult problem to estimate than cumulative cost. If the value function of RL modeling can accurately estimate cumulative conversions, then our method should be able to more easily and accurately estimate cumulative cost and cumulative clicks. Why doesn't our method need to estimate cumulative conversions? Because the optimal bidding formula derived from the dual analysis of LP modeling guarantees the maximization of cumulative conversions, the only thing that needs to be solved in the optimal bidding formula is the value of the dual variable, and the solution of the dual variable only depends on the estimation of cumulative cost and cumulative clicks. This is the clever design of our BiCB method that combines existing mathematical analysis conclusions and statistical models. RL methods usually require the construction of RL training infrastructure and an offline simulator. The operation and maintenance of the offline simulator to align with the online environment is considerable cost. In our method, the estimation of cumulative cost and cumulative clicks is an easy supervised learning problem. We can reuse the CTR estimation and user staying time estimation infrastructure in the advertising system \cite{jiang2019xdl}. And BiCB's online estimation Queries Per Second (QPS) is once every few seconds, which is much lower than the QPS by CTR estimation.

Guo \cite{guo2024generative} proposed an artificial intelligence generative bid (AIGB) method. This method learns the correlation between bid sequence and reward based on historical data, so that the corresponding bid sequence can be generated based on the set reward. This method also adopts the idea of sequence modeling. The application scenario of this work is image advertising, with a control frequency of once every half an hour (48 times a day), that is, the sequence length is 48. In our live advertising scenario, it is necessary to control it once every 10 seconds, 8640 times a day, that is, the sequence length is 8640 times. In practice, the computational load is huge and training is difficult to converge. 

\section{Offline Empirical Evaluation}

\subsection{Dataset}
For offline experiments, we use the public dataset from the NeurIPS 2024 competition "Auto-Bidding in Large-Scale Auctions: Learning
Decision-Making in Uncertain and Competitive Games" \cite{xu2024auto}. The dataset consists of 21 advertising periods. Each period contains over 500,000 impressions. Each period is divided into 48 steps. The advertisers can use the results before step $t$ to refine their strategy for step $t+1$. We use 6 days of data from Official Round-Final Round as the training set and evaluate the result using the last day's data.

With this auction dataset, we evaluate various online bidding algorithms under both BCB (without CPC constraint) and BiCB (with CPC upper and lower bound constraints) settings. In BiCB setting, we generated CPC bounds according to the data distribution provided by advertisers.

\subsection{Metrics}

The objective of the advertisers is to maximize the total conversion under budget and CPC constraints. So we mainly evaluate the performance of the bidding algorithms through the following indicators.

\begin{itemize}
\item Revenue: the cumulative conversion of all advertisers.

\item Cost: the cumulative cost of all advertisers.


\item Budget achievement Rate (BR): Cost/Budget, used to measure budget utilization.

\item $R/R^*$: let $R$ be the revenue of the algorithm, $R^*$ be the optimal revenue with linear programming algorithm. $R/R^*$ measures the distance to the optimal solution.

\item Over-Constrained Ratio (OCR): Number of campaigns that do not satisfy the CPC constraint divided by number of all campaigns. This indicator is only used in BiCB setting.

\item G metric: Following \cite{he2021unified}, we use a G metric to measure the performance under CPC constraints. The metric is defined as:

\begin{equation}
    G=min(\frac{R}{R^*},1.0)-\sum_jp_j
\end{equation}

Where $p_j=\lambda^{exr_j}-1$ represents the penalty for violating constraint $j$. $\lambda\in(1,+\infty)$ is a hyper-parameter. In our experiments, $\lambda$ is set to 100. $exr_j=max(0,CPC/C_u-1)$ for upper bound constraint, and $exr_j=max(0,C_l/CPC-1)$ for lower bound constraint.

\end{itemize}

\subsection{Baselines}

\begin{itemize}
    \item Offline LP: With the supernatural abilities that can predict the traffic details of the whole day, offline linear programming gives the theoretical optimal bidding solution.
    \item BiCB*: BiCB algorithm when the prediction model can make completely accurate predictions of future traffic. 
    \item Manual bidding: Use a fixed bid given by the advertiser for all impressions.
    \item Local PID\cite{yang2019bid}: Use independent PID control system for all dual variables. The dual variables are adjusted in real time according to whether the cost speed or CPC in the most recent time window meets the constraints.
    \item Online LP: Do linear programming on previous training data to get the optimal dual variables, and use this solution to bid in the test set.
    \item IQL\cite{kostrikov2021offline}: A general offline RL approach.
    \item DT\cite{chen2021decision}: An AIGB method. Regard the auto-bidding problem as a generative sequential decision-making problem and solve it with a transformer architecture \cite{han2021transformer}.
\end{itemize}

\subsection{Experimental Results}

We compare the BiCB algorithm with the current mainstream bidding algorithms under two experimental settings. In BCB setting, each approach aims to maximize the total revenue under a fixed budget constraint. In BiCB setting, the algorithms also need to make sure the the cumulative CPC of the entire delivery period meets the upper and lower bound constraints. Detailed data are shown in table \ref{tab:offline_exp_BCB},\ref{tab:offline_exp_BiCB}.

\begin{table}[h]
    \centering
\resizebox{0.47\textwidth}{!} {
    \begin{tabular}{ccccc}
        Method & Revenue & Cost & BR & $R/R^*$ \\
        \toprule
        Manual bid & 1163.21 & 87289.86 & 58.19\% & 57.01\% \\ 
        Local PID & 1308.02 & \textbf{149788.03} & \textbf{99.86\%}  & 64.10\% \\ 
        \midrule
        Online LP & 1594.54 & 146025.57 & 97.35\% & 78.15\% \\ 
        IQL (RL) & 1505.75 & 113165.02 & 75.44\% & 73.79\% \\ 
        DT (AIGB) & 1534.93 & 142705.18 & 95.14\% & 75.22\% \\ 
        BiCB & \textbf{1628.06} & 148090.72 & 98.73\% & \textbf{79.79\%} \\ 
        \midrule
        Offline LP (Optimal) & 2040.47 & 149998.09 & 100.00\% & 100.00\% \\ 
        BiCB* & 2039.23 & 149927.55 & 99.95\% & 99.93\% \\
        \bottomrule
    \end{tabular}
}
    \caption{Cumulative revenue, cost of all approaches under BCB setting.}
    \label{tab:offline_exp_BCB}
\end{table}

\begin{table}[h]
    \centering
\resizebox{0.47\textwidth}{!} {
    \begin{tabular}{ccccccc}
        Method & Revenue & Cost & BR & $R/R^*$ & OCR & G \\
        \toprule
        Manual bid & 286.33 & 48259.00 & 64.35\% & 56.48\% & \textbf{0.00\%} & 0.56\\ 
        Local PID & 297.64 & 49434.05 & 65.91\% & 58.72\% & 20.83\% &  0.54\\ 
        \midrule
        Online LP & 383.80& \textbf{57728.27} & \textbf{76.97\%} & 75.71\%& 35.42\% &  0.57\\ 
        BiCB & \textbf{403.03} & 50043.55 & 66.72\% & \textbf{79.51\%} & 14.58\% & \textbf{0.71}\\ 
        \midrule
        Offline LP (Optimal) & 506.92 & 54457.75 & 72.61\% & 100.00\% & 0.00\% & 1.00 \\ 
        BiCB* & 489.65 & 54491.99 & 72.66\% & 96.59\% & 0.00\% & 0.97 \\
        \bottomrule
    \end{tabular}

}
    \caption{Cumulative revenue, cost of all approaches under BiCB setting.}
    \label{tab:offline_exp_BiCB}
\end{table}

\begin{table*}[t]
    \centering
    \begin{tabular}{cccccccc}
        Method & Budget & Cost & GMV & ROI & CPC & BR & OCR \\
        \toprule
        Local PID & 263238 & 242514 & 2381333 & 9.82 & 1.47 & 92\% & 12.2\% \\
        BiCB & 263238 & 240477(-0.84\%) & 2483205(+4.28\%) & 10.33(+5.16\%) & 1.50(+2.50\%) & 91\%(-0.84\%) & 11.4\%(-6.25\%) \\ 
        \bottomrule
    \end{tabular}
    \caption{BiCB online experiment results.}
    \label{tab:online_exp}
\end{table*}
\subsubsection{BiCB vs state-of-the-art approaches}

As shown in table \ref{tab:offline_exp_BCB}, under BCB setting, BiCB algorithm outperforms traditional algorithms like local PID. Besides, comparing with Online LP, RL and AIGB methods, our BiCB algorithm achieves similar or even better results. Since we use a lightweight model, our computational cost is much lower than these methods, which allows us to deploy on a second-level control system for live advertising.

Under BiCB experimental setting, as shown in table \ref{tab:offline_exp_BiCB}, traditional bidding algorithms perform poorly and can only approach less than 60\% of the optimal solution. The Online LP algorithm achieved high revenue, but due to the difference in the distribution of historical data and today's data, 35\% of the advertisers do not satisfy the CPC constraint. And our lightweight approach outperforms other methods on CPC constraint satisfaction, total revenue, and G metric.

\subsubsection{Offline LP vs BiCB* and BiCB}

Our BiCB with the trained LightGBM model achieves 80\% of the optimal solution. But when the offline prediction model can accurately predict future traffic, the BiCB algorithm can achieve more than 96\% of the theoretical optimal solution. This proves that our greedy algorithm has a very high theoretical upper limit as the estimation accuracy improves.

\subsubsection{Computional Complexity}
We compare the computational performance of BiCB and AIGB methods in offline experiments. Under the setting that live ads are adjusted every 10 seconds, on a single CPU, RL methods can support the control of 200 orders. In contrast, the LightGBM model achieves up to 80k inference QPS, and enabling control of 1k orders, which effectively reduces computational complexity.

\section{Online Empirical Evaluation}

\subsection{Experimental Setup}

Since live streaming advertising has a very high demand for timeliness, it is necessary to quickly adapt to changes in the live streaming rhythm through offline second-level regulation. Algorithms such as Online LP, RL and AIGB have poor adaptability to live streaming due to their heavy models and long decision-making time, so they do not perform well online. Therefore, our current baseline algorithm still uses the local PID algorithm.

Currently, the BiCB algorithm has been fully deployed in production of Alibaba's live streaming advertising. We compare BiCB with online baselines through budget bucketing experiments. The experiment includes totally 263 campaigns. Each advertising campaign is divided into four equal parts according to the budget, and two of them are used as the baseline and control group respectively. The experimental period is from February 5th to 8th. 

\subsection{Evaluation Results}

The results of BiCB online experiment are shown in the table \ref{tab:online_exp}. Comparing with baseline local PID, our BiCB algorithm gains 4.28\% increase in GMV. And the number of orders that do not meet CPC constraints has decreased by 6.25\%. The results show that the BiCB algorithm can create higher transaction value for advertisers.

\subsection{Analysis}

\subsubsection{Stability}

\begin{figure}[h]
    \centering
    \begin{subfigure}{0.49\linewidth}
    \includegraphics[width=\linewidth]{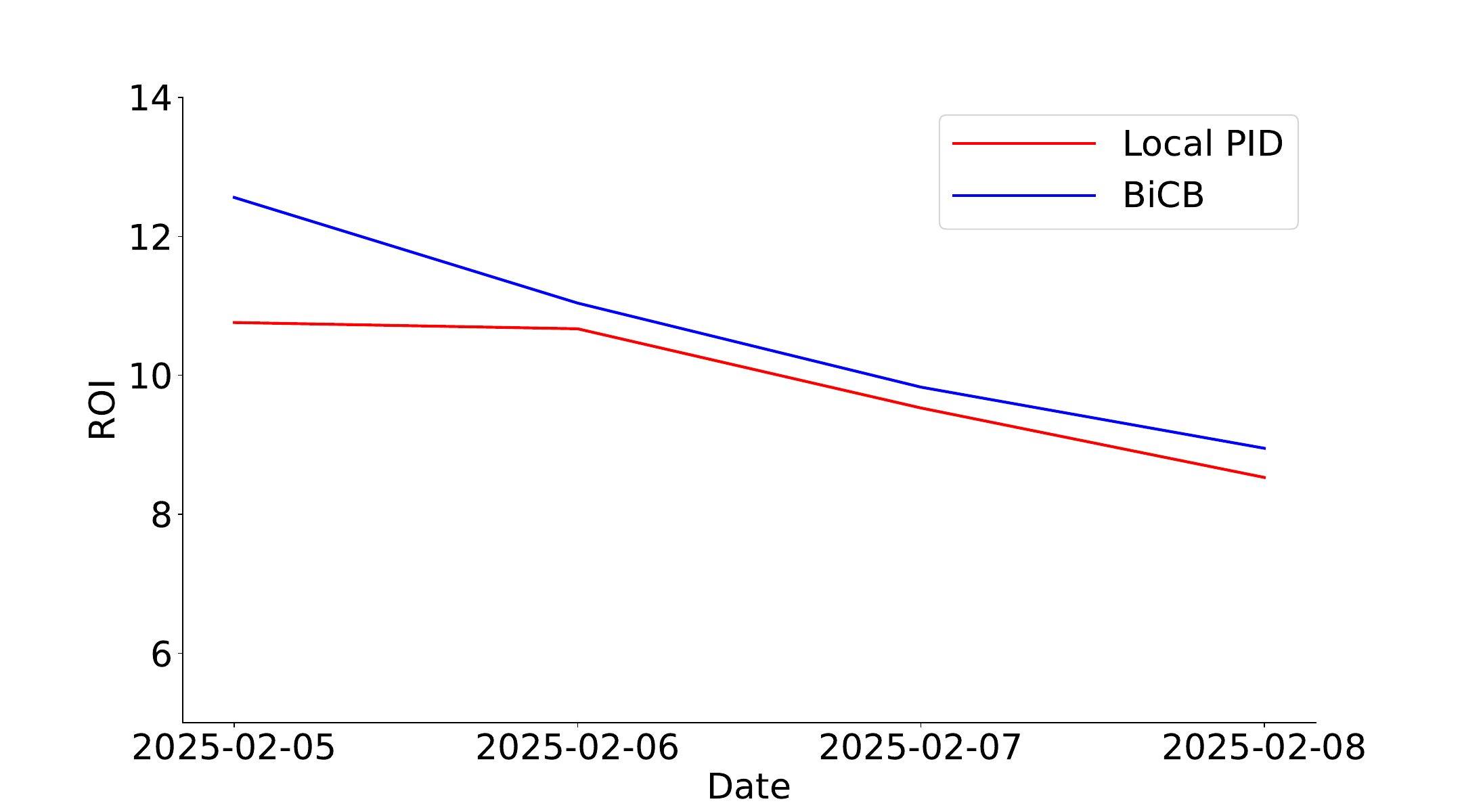}
    \end{subfigure}
    \begin{subfigure}{0.49\linewidth}
    \includegraphics[width=\linewidth]{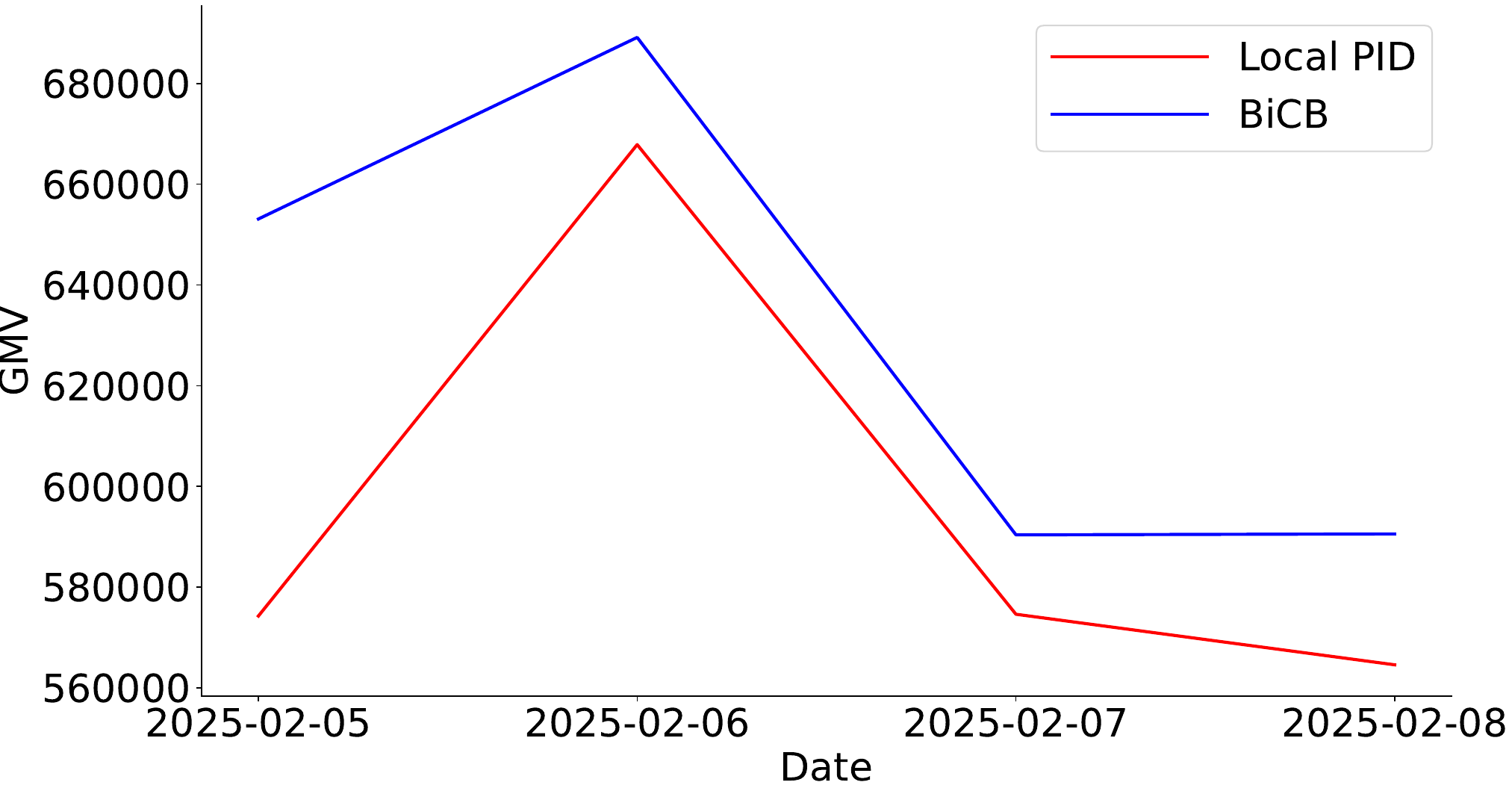}
    \end{subfigure}
    \caption{The ROI and GMV curve of online experiment for Local PID and BiCB.}
    \label{fig:roi_gmv_day}
\end{figure}

Figure \ref{fig:roi_gmv_day} shows the ROI and GMV curves of the online algorithm every day during the experimental period. The red line is the local PID algorithm, and the blue line is the BiCB algorithm. It can be seen that our method has a stable positive improvement compared to the online baseline method.

\subsubsection{Case Study}

\begin{figure}[h]
    \centering
    \begin{subfigure}{0.49\linewidth}
        \includegraphics[width=\linewidth]{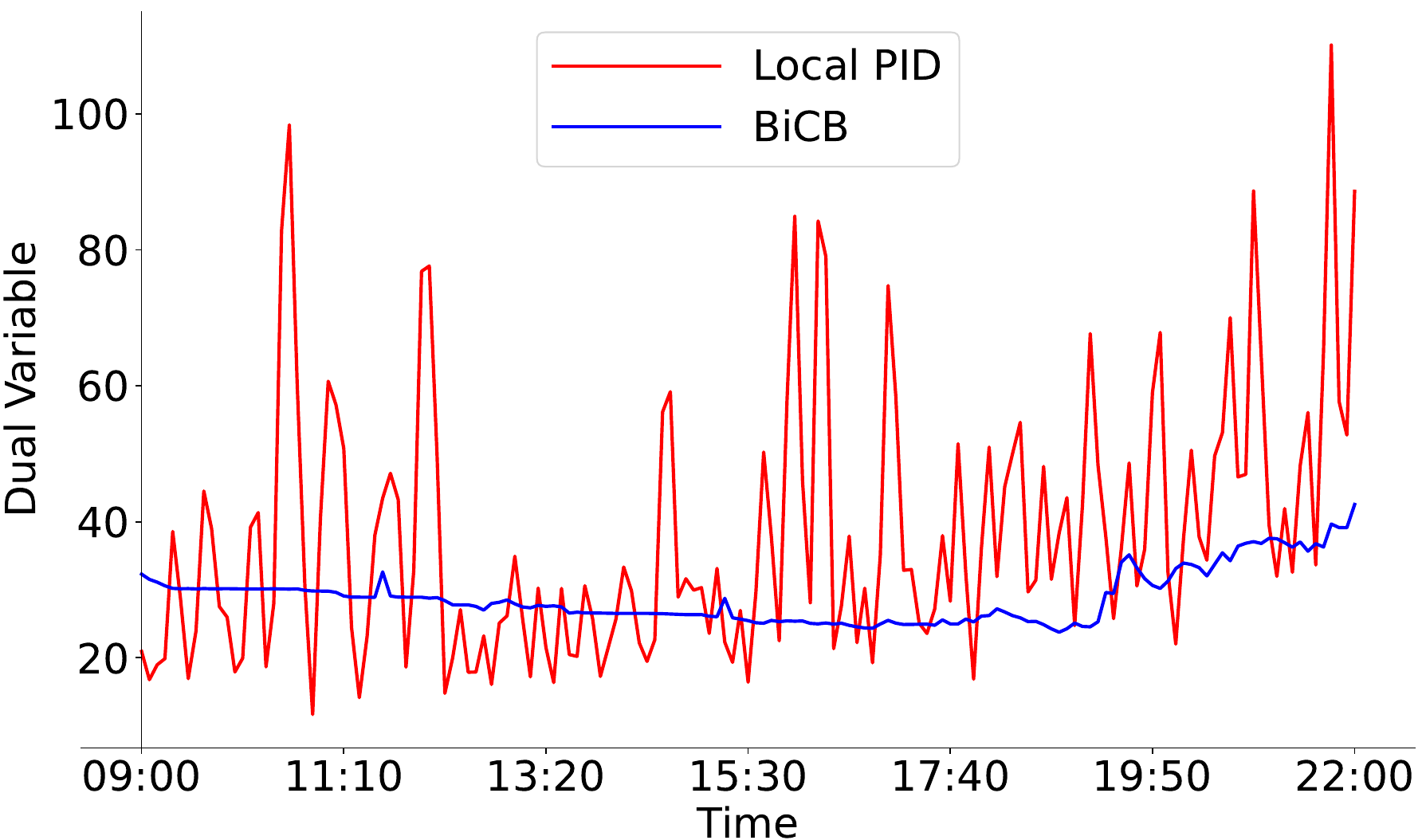}
        \label{fig:p_time}
    \end{subfigure}
    \begin{subfigure}{0.49\linewidth}
        \centering
    \includegraphics[width=\linewidth]{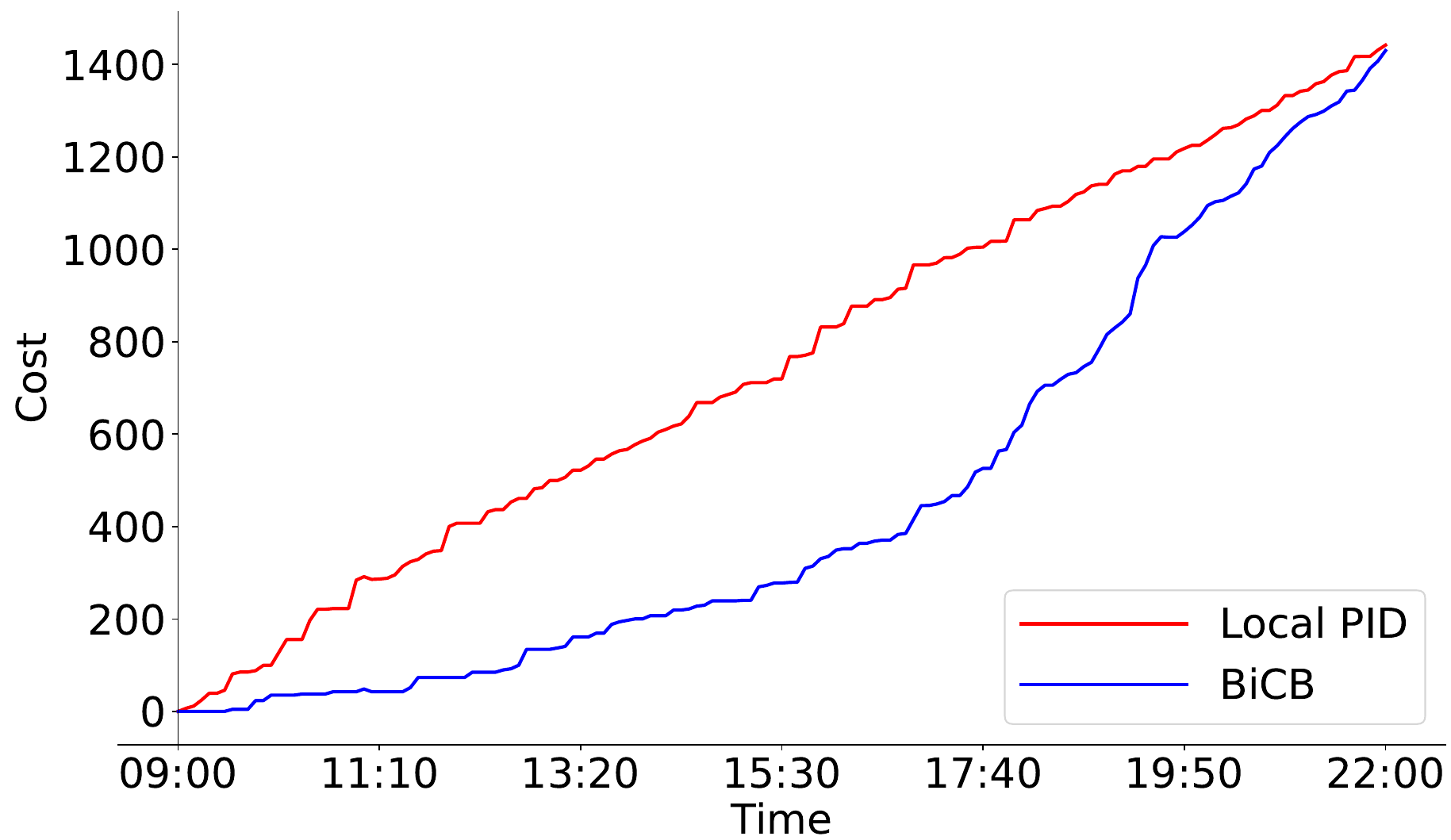}
    \label{fig:cost_time}
    \end{subfigure}
    \caption{Dual variable and cumulative cost of Local PID and BiCB algorithms over time. The Local PID algorithm shows uniform cost but unstable bidding parameters over time, while the BiCB algorithm has more stable bidding parameters, and allocates more budget to the more efficient evening hours, thus achieving higher efficiency.}
    \label{fig:p_cost_time}
\end{figure}

In order to more intuitively reflect the advantages of the BiCB algorithm, Figure 
\ref{fig:p_cost_time} plots the changes in the dual variables and cumulative cost of an online order within a day.

The red line is the baseline local PID algorithm, and the blue line is the BiCB algorithm. Since the local PID only observes the delivery data of the adjacent time window, it achieves uniform cost throughout the day, but the bidding parameters fluctuate violently. The BiCB algorithm takes into account the estimation of future traffic, so it is not consumed uniformly, and the signal is very stable within a day with little fluctuation. Stable bidding parameters ensure that the algorithm has a constant value measurement throughout the day, thereby achieving better delivery results. From the perspective of mathematical modeling, these parameters are dual variables, and only when they maintain constant values over all traffic requests can the optimality of the solution be guaranteed.

    

\section{Conclusion}
Based on the scenario where live advertising requires high-frequency bidding control, we have improved the linear programming modeling using the upper and lower bound constraints of auto-bidding. We have given a solution to this problem, namely BiCB algorithm, a lightweight auto-bidding algorithm. BiCB is based on the optimal bidding formula of LP dual analysis, through a lightweight future traffic prediction module, and adopts an approximate algorithm with low complexity to achieve excellent optimal solution approximation performance. We have given theoretical analysis of the approximation of the optimal solution of BiCB algorithm and introduced the engineering implementation of the algorithm in the live advertising scenario. We have also discussed the technical points of the mainstream practical solutions for auto-bidding in the industry. Sufficient offline experiments revealed the advantage of BiCB. The algorithm has been fully deployed in the live advertising business of a large Internet platform and has achieved positive business results.
\begin{acks}
We would like to thank Dongdong Peng and Fanduo Li for their invaluable help in data cleaning and plotting.
\end{acks}

\bibliographystyle{ACM-Reference-Format}
\bibliography{sample-base}


\appendix

\section*{APPENDIX}

\section{Proof for Theorem \ref{thm1}}


We first prove a lemma:

\newtheorem{lem1}{Lemma}

\begin{lem1}
\label{lem1}
$\forall \epsilon>0$, there exists a sequence $\{\epsilon_i\},0\le \epsilon_i<\epsilon$, such that for any three distinct points $(wp_i,obj_i+\epsilon_i),(wp_j,obj_j+\epsilon_j),(wp_k,obj_k+\epsilon_k),i<j<k$, they cannot be collinear.
\end{lem1}

Proof for lemma \ref{lem1}:

Consider the construction method of $\{\epsilon_i\}$:
Let $\epsilon_0=0$, for all $k>0$, enumerate all $i,j$ that satisfy $0\le i<j<k$. Obviously, for each pair $i,j$, there exists at most one $\hat{\epsilon_{ijk}}$ to make $(wp_i,obj_i+\epsilon_i),(wp_j,obj_j+\epsilon_j),(wp_k,obj_k+\hat{\epsilon_{ijk}})$ collinear. And there are infinite values for $\epsilon_k \in [0,\epsilon)$. So $[0,\epsilon) \backslash \{\hat{\epsilon_{ijk}}\}$ is not empty. Let $\epsilon_k$ be any value in this set. The constructed $\{\epsilon_i\}$ can guarantee that any three points are not collinear.

For any feasible solution to problem $(wp, obj, B, C_u, C_l)$, it must also be a feasible solution to problem $(wp, \{obj_i+\epsilon_i\}, B, C_u, C_l)$ (because the constraints are independent of obj).

And since ${C_u\sum_i x_i*pctr_i*obj_i}\le{C_u\sum_i x_i*pctr_i*(obj_i+\epsilon_i)}\\\le(1+\frac{\epsilon}{\mathop{min}\limits_i obj_i}){C_u\sum_i x_i*pctr_i*obj_i}$, the lemma holds.

This lemma proves that a small perturbation for $obj$ of the original problem can ensure that any three $(wp_i, obj_i)$ are not collinear and are close enough to the optimal solution of the original problem. Therefore, we can only discuss the situation that any three $(wp_i, obj_i)$ are not collinear.

Get back to theorem \ref{thm1}, from the derivation of the linear programming problem and the definition of the greedy solution, when $obj_i\ne \frac{wp_i*(p^*+q_u^*-q_l^*)-C_u*q_u^*+C_l*q_l^*}{C_u}$, $x_i$ in greedy solution and optimal solution solution are equivalent. And from the lemma \ref{lem1}, there are at most 2 points that satisfies $obj_i= \frac{wp_i*(p^*+q_u^*-q_l^*)-C_u*q_u^*+C_l*q_l^*}{C_u}$. Thus the error between the greedy solution and optimal solution solution is at most $2\mathop{max}\limits_i obj_i$, and the impact on CPC is at most two clicks. So $CPC_d^G \le B/(\frac{B}{C_u}-2)=C_u+\frac{2C_u^2}{B-2C_u}$ and $CPC_d^G \ge (B-2\mathop{max}\limits_i wp_i)/(\frac{B}{C_l})=C_l-\frac{2C_l\mathop{max}\limits_i wp_i}{B}$. Therefore, $CPC_d^G \in(C_u+\epsilon^c,C_l-\epsilon^c)$.

Then we can get $V_d^G(wp,obj,B,C_u+\epsilon^c,C_l-\epsilon^c)\ge \\V_c^*(wp,obj,B,C_u,C_l)-2\mathop{max}\limits_i obj_i$. Obviously, $V_c^* \ge V_d^*$(because linear programming problem has a larger solution space), so $\\V_d^G(wp,obj,B,C_u+\epsilon^c,C_l-\epsilon^c)\ge V_d^*(wp,obj,B,C_u,C_l)-2\mathop{max}\limits_i obj_i$ holds.

\section{Convergence of Projected Gradient Descent}

In linear programming formulation, each impression is a discrete point, so the cost and clicks are not continuous functions about $p,q_u,q_l$. For convenience, we convert the original problem into a continuous problem, and define the cost and clicks when the dual variables are $p,q_u,q_l$ as the following functions:

\begin{align}
cost(p,q_u,q_l)=\int_{0}^{+\infty}\int_{\frac{w*(p+q_u-q_l)-C_u*q_u+C_l*q_l}{C_u}}^{+\infty}wF(w,v)dvdw \\
click(p,q_u,q_l)=\int_{0}^{+\infty}\int_{\frac{w*(p+q_u-q_l)-C_u*q_u+C_l*q_l}{C_u}}^{+\infty}F(w,v)dwdv
\end{align}

$F$ represents the impression distribution, e.g. $F(w,v)=\\\frac{\lambda}{\pi}\sum_{i=1}^n pctr_i*e^{-\lambda[(w-wp_i)^2+(v-obj_i)^2]}$, when $\lambda\rightarrow\infty$, the cost and clicks calculated by this continuous functionare equivalent to the original discrete problem.

According to the complementary relaxation principle, $q_u>0,q_l>0$ cannot be true at the same time. To simplify the notation, we will use a unified variable $q$ to represent $q_u,q_l$ in the following text. $cost(p,q)$ is defined as $cost(p,q)=cost(p,q,0)$ when $q_u\ge 0$, otherwise $cost(p,q)=cost(p,0,-q)$. $click(p,q)$ is defined in the same way.

We can get

\begin{align}
\nonumber
    \frac{\partial cost}{\partial p}&=\int_{0}^{+\infty}wF(w,\frac{w*(p+q_u-q_l)-C_u*q_u+C_l*q_l}{C_u}) \\
    &*(-\frac{w}{C_u})dw<0
\end{align}

which means $cost(p,q)$ is monotonically decreasing with respect to $p$. Define $P(q)=if~cost(0,q)<B~then~0~else~p|cost(p,q)=B$. From the monotonicity, we can see the definiton is self-consistent.

\begin{thm1}
\label{thm2}
When $q\ge 0$, $cost(P(q),q)-C_u*click(P(q),q)$ is monotonically decreasing with respect to q. When $q<0$, $cost(P(q),q)-C_l*click(P(q),q)$ is monotonically decreasing with respect to q. 
\end{thm1}

Proof:

When $q\ge 0$, we have

\begin{align}
\frac{\partial cost}{\partial p}&=-\int_{0}^{+\infty}w^2F(w,\theta_0(w))/C_udw \\
\frac{\partial cost}{\partial q}&=-\int_{0}^{+\infty}w(w-C_u)F(w,\theta_0(w))/C_udw \\
\frac{\partial click}{\partial p}&=-\int_{0}^{+\infty}wF(w,\theta_0(w))/C_udw \\
\frac{\partial click}{\partial q}&=-\int_{0}^{+\infty}(w-C_u)F(w,\theta_0(w))/C_udw \\
\end{align}

where $\theta_0(w)=\frac{w*(p+q)-C_u*q}{C_u}$.

So that

\begin{align}
    &\frac{\partial[cost(P(q),q)-C_u*click(P(q),q)]}{\partial q} \\ 
&=\frac{\partial (cost-C_u*click)}{\partial P}\frac{dP}{dq}+\frac{\partial (cost-C_u*click)}{\partial q} \\ 
&=\frac{\partial (cost-C_u*click)}{\partial p}(-\frac{\partial cost}{\partial q}/\frac{\partial cost}{\partial p})+\frac{\partial (cost-C_u*click)}{\partial q} \\
&= C_u*(\frac{\partial click}{\partial p}*\frac{\partial cost}{\partial q}-\frac{\partial cost}{\partial p}*\frac{\partial click}{\partial q})/\frac{\partial cost}{\partial p}
\\
\nonumber
&=C_u(\int_{0}^{+\infty}w^2F(w,\theta_0(w)))dw*\int_{0}^{+\infty}F(w,\theta_0(w)))dw- \\
&(\int_{0}^{+\infty}wF(w,\theta_0(w)))dw)^2)/\frac{\partial cost}{\partial p}
\end{align}

By Cauchy inequality, 

$\int_{0}^{+\infty}w^2F(w,\theta_0(w)))dw*\int_{0}^{+\infty}F(w,\theta_0(w)))dw \\ \ge (\int_{0}^{+\infty}wF(w,\theta_0(w)))dw)^2$, and $\frac{\partial cost}{\partial p}<0$

so we can get \\ $\frac{\partial[cost(P(q),q)-C_u*click(P(q),q)]}{\partial q}\le 0$.

Similarly, when $q<0$, we can infer that




\begin{align}
&\frac{\partial[cost(P(q),q)-C_l*click(P(q),q)]}{\partial q} \\ 
\nonumber
&=\frac{C_l^2}{C_u}(\int_{0}^{+\infty}w^2F(w,\theta_1(w)))dw*\int_{0}^{+\infty}F(w,\theta_1(w)))dw- \\
&(\int_{0}^{+\infty}wF(w,\theta_1(w)))dw)^2)/\frac{\partial cost}{\partial p}
\end{align}

where $\theta_1(w)=\frac{w*(p+q)-C_l*q}{C_u}$.

With Cauchy inequality and $\frac{\partial cost}{\partial p}<0$, we can infer that 

$\frac{\partial[cost(P(q),q)-C_u*click(P(q),q)]}{\partial q}\le 0$. So the theorem is proved.

\begin{thm1}
\label{thm3}
For the sequence

\begin{align}
q_u^{t+1}&=q_u^t+\gamma^t*(cost(p^t,q_u^t,q_l^t)-C_u*click(p_t,q_u^t,q_l^t)) \\
q_l^{t+1}&=q_l^t+\theta^t*(cost(p^t,q_u^t,q_l^t)-C_l*click(p_t,q_u^t,q_l^t)) \\
p^{t+1}&=p^t+\lambda^t*(cost(p^t,q_u^{t+1},q_l^{t+1})-B)
\end{align}

$\exists~\{\lambda^t\},\{\gamma^t\},\{\theta^t\}, \lambda^t,\gamma^t,\theta^t\ge0$, so that $\{p^t\},\{q_u^t\},\{q_l^t\}$ converge, and the convergent point satisfies equation \ref{eq:relax_B}, \ref{eq:relax_cu}, \ref{eq:relax_cl}.

\end{thm1}

Define $q^t=if~q_u^t>0~then~q_u^t~else~q_l^t$, let $q_u^0=q_l^0=0$, $p^0=P(0)$

If the initial value satisfies both CPC upper and lower bound constraints, then it is already optimal solution. If it does not satisfy the upper bound constraint, then the optimal solution satisfies $q_u^*>0,q_l^*=0$. If it does not satisfy the lower bound constraint, then the optimal solution satisfies $q_u^*=0,q_l^*>0$.

Take the example of not satisfying the upper bound constraint, consider the following construction method.

Since the optimal solution satisfies $q_l^*=0$, so let $\theta^t$ be always $0$.

Let $M$ be a large enough constant, $l^0=r^0=M$.

If $r^{t-1}>\epsilon$, then 

\begin{align}
    l^t&=l^{t-1} \\
    r^t&=r^{t-1}/2 \\
    \gamma^t&=0 \\
    \lambda^t&=\frac{r^{t-1}}{|cost(p^{t-1},q^{t-1})-B|}
\end{align}

otherwise,

\begin{align}
    l^t&=l^{t-1}/2 \\
    r^t&=M \\
    \gamma^t&=\frac{l_t}{|cost(p^{t-1},q^{t-1})-C_u*click(p^{t-1},q^{t-1})|} \\
    \lambda^t&=0
\end{align}

This procedure is equivalent to a binary search procedure. And according to the monotonicity in theorem \ref{thm2}, it will converge to the optimal solution.

For $\forall \epsilon>0$, it takes $O(log\frac{1}{\epsilon})$ steps for $p$ to converge, and $O(log\frac{1}{\epsilon})$ steps for $q$ to converge. So totally it needs $O((log\frac{1}{\epsilon})^2)$ iterations to converge.

\section{Proof for $p^*+q_u^*-q_l^*>0$ in BiCB}

In real-world scenarios, since the objective function is often positively correlated to all advertisers, $\{wp_i\}$ and $\{obj_i\}$ are often strongly positively correlated from a single customer perspective. We make the following assumption about real-world scenarios:

\newtheorem{assu1}{Assumption}

\begin{assu1}
\label{assu1}
\begin{align}
&\exists obj_{thres}>0 \\
s.t.&\int_{0}^{+\infty}\int_{obj_{thres}}^{+\infty}wF(w,v)dvdw=B \\
&\frac{\int_{0}^{+\infty}\int_{obj_{thres}}^{+\infty}wF(w,v)dvdw}{\int_{0}^{+\infty}\int_{obj_{thres}}^{+\infty}F(w,v)dvdw} > C_l
\end{align}

\end{assu1}

That is for the impression subset with the highest objective, the total CPC is higher than $C_l$.

\begin{thm1}
\label{thm4}
Under Assumption \ref{assu1}, $p^*+q_u^*-q_l^*>0$

\end{thm1}

When $q_l^*=0$, $p^*+q_u^*-q_l^*>0$ is obvious. So we only need to discuss the case where $q_u=0,q_l>0$.

From Assumption \ref{assu1}, when $q_u=0,q_l=\frac{C_u}{C_l}obj_{thres},q=-q_l$, we have

$cost(q_l,0,q_l)=\int_{0}^{+\infty}\int_{obj_{thres}}^{+\infty}wF(w,v)dvdw=B$

So we can infer 

\begin{align}
    &P(q)=q_l \label{eq:pq}\\
    &cost(P(q),q)-C_l*click(P(q),q)>0
\end{align}

According to equation \ref{eq:relax_cl}, the optimal solution satisfies \\$cost(P(q^*),q^*)-C_l*click(P(q^*),q^*)=0$.

According to theorem \ref{thm2}, $cost(P(q),q)-C_l*click(P(q),q)$ decreases monotonically with respect to $q$. So that 

\begin{align}
    q^*>q \\
    q_l^*<q_l \label{eq:uneq}
\end{align}

Since

\begin{align}
&\frac{d(P(q)+q)}{dq} \\
&=-\frac{\partial cost}{\partial q}/\frac{\partial cost}{\partial p}+1 \\
&=-\frac{C_l}{C_u}\int_{0}^{+\infty}wF(w,\theta_1(w)dw/\frac{\partial cost}{\partial p} >0
\end{align}

We have 

\begin{equation}
    d(P(-q_l)-q_l)/dq_l<0
    \label{eq:der}
\end{equation}

Combining (\ref{eq:pq}), (\ref{eq:uneq}), (\ref{eq:der}), we can infer that $P(-q_l^*)-q_l^*=p^*+q_u^*-q_l^*>0$, so the theorem is proved.

\end{document}